\documentclass[sn-basic,Numbered,iicol]{sn-jnl}

\usepackage{CJKutf8}

\usepackage{graphicx}%
\usepackage{multirow}%
\usepackage{multicol}%
\usepackage{amsmath,amssymb,amsfonts}%
\usepackage{amsthm}%
\usepackage{mathrsfs}%
\usepackage[title]{appendix}%
\usepackage{xcolor}%
\usepackage{textcomp}%
\usepackage{manyfoot}%
\usepackage{booktabs}%
\usepackage{algorithm}%
\usepackage{algorithmicx}%
\usepackage{algpseudocode}%
\usepackage{listings}%
\usepackage{caption,subcaption}

\theoremstyle{thmstyleone}%
\theoremstyle{thmstyletwo}%

\theoremstyle{thmstylethree}%

\begin{document}

\title[Article Title]{BatGPT-Chem: A Foundation Large Model For Retrosynthesis Prediction}


\author[1]{Yifei Yang}
\equalcont{These authors contributed equally to this work.}
\author[1]{Runhan Shi}
\equalcont{These authors contributed equally to this work.}
\author[2]{Zuchao Li}
\author[3]{Shu Jiang}
\author[1]{Bao-Liang Lu}
\author[1]{Yang Yang$^*$}\email{yangyang@cs.sjtu.edu.cn}
\author[1]{Hai Zhao$^*$}\email{zhaohai@cs.sjtu.edu.cn}

\affil[1]{Department of Computer Science and Engineering, and Key Laboratory of Shanghai Education Commission for Intelligent Interaction and Cognitive Engineering, Shanghai Jiao Tong University, 800 Dongchuan Road, Shanghai 200240, China}

\affil[2]{Wuhan University, China}

\affil[3]{Nantong University, China}


\abstract{
Retrosynthesis analysis is pivotal yet challenging in drug discovery and organic chemistry. Despite the proliferation of computational tools over the past decade, AI-based systems often fall short in generalizing across diverse reaction types and exploring alternative synthetic pathways. This paper presents BatGPT-Chem, a large language model with 15 billion parameters, tailored for enhanced retrosynthesis prediction. Integrating chemical tasks via a unified framework of natural language and SMILES notation, this approach synthesizes extensive instructional data from an expansive chemical database. Employing both autoregressive and bidirectional training techniques across over one hundred million instances, BatGPT-Chem captures a broad spectrum of chemical knowledge, enabling precise prediction of reaction conditions and exhibiting strong zero-shot capabilities. Superior to existing AI methods, our model demonstrates significant advancements in generating effective strategies for complex molecules, as validated by stringent benchmark tests. BatGPT-Chem not only boosts the efficiency and creativity of retrosynthetic analysis but also establishes a new standard for computational tools in synthetic design. This development empowers chemists to adeptly address the synthesis of novel compounds, potentially expediting the innovation cycle in drug manufacturing and materials science. We release our trial platform at \url{https://www.batgpt.net/dapp/chem}.
}



\keywords{retrosynthesis, LLMs, chemical reactions}



\maketitle

\section{Introduction}\label{sec1}
Retrosynthesis analysis~\cite{retro_0}, which aims to identify a set of precursors given a target molecule, is essential in synthetic chemistry. It plays a vital role in applications like drug design, chemical biology, and material science. However, the extensive range of possible chemical transformations and the incomplete understanding of the chemical reaction mechanisms make retrosynthesis planning an extremely challenging task, even for experienced chemists addressing smaller molecular structures. 
Over recent decades, the development of various computer-aided synthesis planning (CASP) methods has emerged to address these challenges~\cite{retro_0, retro_1, retro_2}. Artificial intelligence (AI)-based methods, specifically in reaction modeling frameworks, combined with the increasing availability of extensive synthetic datasets, has facilitated the advancement of data-driven approaches, notably deep learning (DL) models. 
These approaches have significantly aided chemists, saving considerable time and effort in designing synthetic experiments. 

Data-driven retrosynthesis methodologies can be broadly categorized into three types: template-based, template-free, and semi-template-based. A reaction template is essentially a subgraph pattern that illustrates the changes in atoms and bonds between a product molecule and its reactants. Template-based approaches~\cite{TB_0, TB_1, TB_2, TB_3} incorporate fundamental reactive rules that are either manually defined or derived from chemical reaction datasets, enabling them to delineate molecular transformations during reactions. Typically, these algorithms rely on a repository of reaction templates used to align target molecules and transform product molecules into reactants through pertinent templates. Although template-based methods offer high interpretability and produce chemically coherent reactants with perfect validity, their scope and complexity remain constrained by the limited ability to generate reactions beyond the template library.

In contrast, both template-free and semi-template methods are independent of an external template database, thereby enhancing model generalization capabilities. Template-free methods~\cite{TF_0, TF_1, TF_2, TF_3, TF_4, TF_5, TF_6, TF_8, TF_9} treat retrosynthesis as a sequential generation problem, transforming products into potential precursors in an end-to-end fashion. Semi-template approaches~\cite{STB_0, STB_1, STB_2, STB_3, STB_4} partition retrosynthesis into two phases: initially identifying the reaction center, i.e., a focal area in a product, to generate intermediate molecules known as synthons, followed by augmenting these synthons to form precursors. Non-template-based methods typically adopt either sequence-based or graph-based approaches. Sequence-based strategies view retrosynthesis as a neural machine translation task, depicting molecules as linear texts, such as simplified molecular input line entry system (SMILES) strings~\cite{smiles}, and utilizing neural language models like Transformer~\cite{Vaswani2017AttentionIA} for prediction. Graph-based approaches, on the other hand, interpret retrosynthesis through molecular graph structures, deploying models such as directed message passing neural networks to make predictions. Compared to template-based methods, 

Leveraging AI technologies, template-free approaches have demonstrated enhanced potential in both generalization and predictive diversity. This research delves into these methods in particular. While AI-driven methods have made considerable progress in retrosynthesis analysis, they still face certain limitations, as outlined below.

i) Deficiency in molecular and chemical reaction knowledge. Traditional AI models, constrained by their learning capacities and limited training data, often fail to incorporate comprehensive knowledge from chemical literature. This significantly limits performance enhancements. Although recent studies have applied advanced Large Language Models (LLMs) to chemistry~\cite{boiko2023autonomous,qian2023can,li2023empowering} to mitigate this limitation, these models have not been adequately fine-tuned on specific chemical datasets.

ii) Neglect of reaction conditions\footnote{In the context of our paper, we use the term ``reaction conditions'' to refer to substrates like solvents and catalysts that do not contribute any atoms to the product. Reaction conditions are not considered as reactants in the context of our paper.}. Current models often exclude substances not directly involved in reactions or arbitrarily mix them up with reactants, due to the lack of reaction conditions in training data and difficulties integrating them. This reduction in interpretability and reliability undermines model performance. Yet, chemical reaction success profoundly depends on precise reaction conditions~\cite{condition_importance} including catalysts, solvents, and other factors that influence reaction rates, yields, and selectivity. Accurate modeling of these conditions is crucial for effective retrosynthesis planning.

iii) Limited zero-shot prediction capability. Typical AI models are trained and tested on data from the same distribution, rendering them ineffective for out-of-distribution predictions, particularly in zero-shot retrosynthesis tasks. This limitation severely restricts their ability to transfer knowledge across different datasets or chemical reaction classes. The problem is particularly pronounced when attempting retrosynthesis across a diverse chemical space~\cite{rea_ELN_BH}.

To enhance the accuracy, reliability, and generalization of retrosynthetic models and thereby improve their utility in chemical synthesis, we develop the innovative BatGPT-Chem for one-step retrosynthesis prediction. In this work, we leverage the widely-used SMILES notation as a specialized chemical language, integrating it with natural language through the use of Large Language Models (LLMs). By combining open-source and closed-source datasets into a larger-scale instruction-tuning dataset, utilizing prompt templates, we effectively facilitate instruction-tuning for BatGPT-Chem.

Significantly, our prompts explicitly include reaction conditions, covering a vast chemical space from various downstream datasets. Building upon our previously developed LLM, BatGPT-15B~\cite{li2023batgpt}, we expand the model's vocabulary with specialized chemical terms and further refined it through instruction tuning, culminating in the model of BatGPT-Chem. This model surpasses existing large chemical models~\cite{zhang2024chemllm,zhao2024chemdfm} in terms of model size and uses a more comprehensive bilingual (Chinese and English) instruction-tuning dataset.

Experimental results validate that BatGPT-Chem successfully incorporates chemical knowl-
edge into retrosynthesis prediction, performs well under zero-shot conditions, and explicitly predicts reaction conditions in an end-to-end manner. This extensive capability sets new benchmarks for the application of LLMs in the field of chemical engineering.

In summary, this study demonstrates the potential of integrating specialized chemical language with advanced instruction-tuning techniques via a powerful LLM to enhance the accuracy and robustness of retrosynthesis analysis. By learning from large-scale datasets of chemical literature and symbolic strings, along with mastering an enriched chemical-specific vocabulary, BatGPT-Chem interprets the relationship between chemical text descriptions and molecules, which is evidenced by the model's capability to predict reaction conditions not included in the reaction routes for training. Such advancements emphasize the profound impact that specialized instructional methodologies can have on the evolution of computational chemistry. Moreover, the deployment of BatGPT-Chem on an online server facilitates chemists’ access to sophisticated suggestions for synthesizing novel compounds, potentially expediting the innovation cycle in drug manufacturing and materials science.

\section{Results}
\subsection{Retrosynthesis prediction benchmark}\label{subsec:data}
Given a set of products, the objective is to generate precursors that synthesize the products. Similar to other LLMs, BatGPT-Chem has been trained on extensive data, making it challenging to estimate its zero-shot prediction ability for retrosynthesis planning. To minimize overlap between test and training data and to explore a broader chemical space, we collect and organize eight datasets with various reaction types to establish a new benchmark dataset for retrosynthesis prediction. We take reaction conditions (precursors) into account as much as possible. Details of the benchmark are provided below.
\begin{figure*}[h]
\centering
    \includegraphics[width=0.8\linewidth]{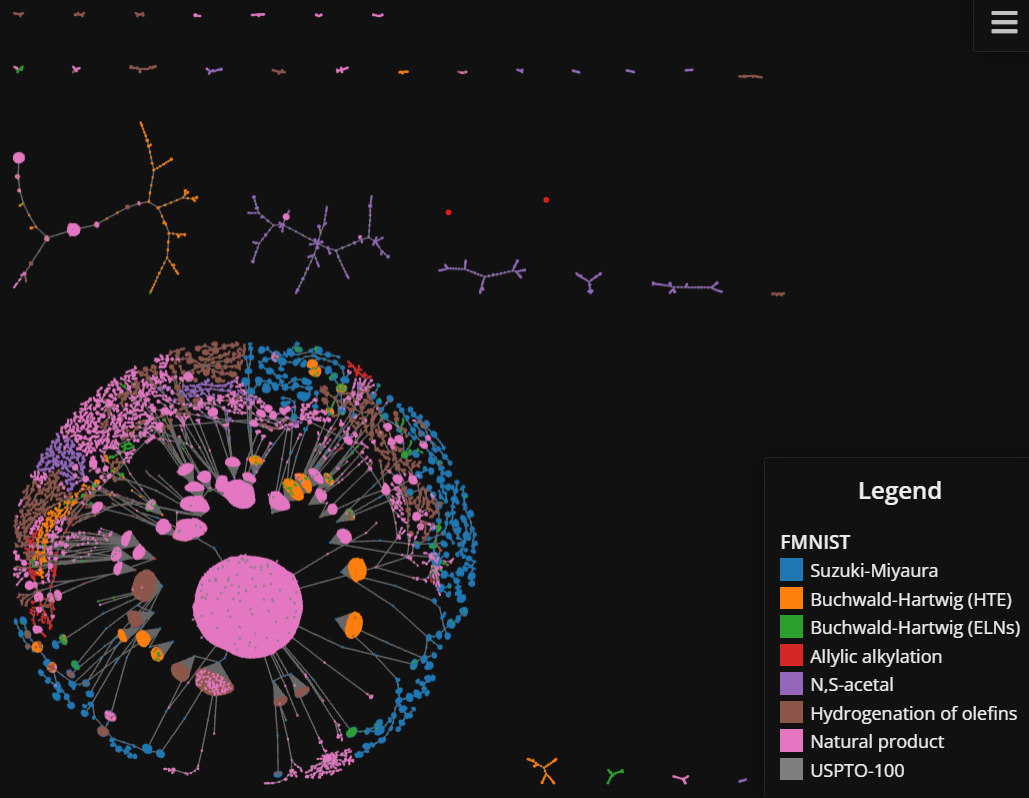}
    \caption{The annotated reaction graphs. The different fingerprints of reactions are visualized using a TMAP algorithm \cite{tmap} and the Faerun visualization library \cite{faerun}.}
    \label{fig:tmap_0}
\end{figure*}
\begin{itemize}
    \item The Suzuki-Miyaura (SM) dataset~\cite{rea_SM} contains one product of the Suzuki-Miyaura cross-coupling reactions. There are 5,760 reactions on the combinations of 15 couplings of electrophiles and nucleophiles, 12 ligands (with a blank one), eight bases (with a blank one), and four solvents.
    \item The high-throughput experiments Buchwald-Hartwig (HTE BH) dataset~\cite{rea_HTE_BH} contains five products of the Pd-catalyzed Buchwald-Hartwig C-N cross-coupling reactions. There are 3,955 reactions on the combinations of 15 aryl halides, four ligands, three bases, and 23 additives.
    \item The electronic laboratory notebooks Buchwald-Hartwig (ELN BH) dataset~\cite{rea_ELN_BH} contains 454 products of the Pd-catalyzed Buchwald-Hartwig C-N cross-coupling reactions. It has 551 reactions with a wider range of reaction space than the HTE BH dataset.
    \item The asymmetric allylic alkylation with amine (AAAA) dataset~\cite{rea_AAAA} contains 189 products of 273 reactions.
    \item The Denmark dataset~\cite{rea_Denmark} contains 25 products of the asymmetric $N,S$-acetal formation using CPA catalysts. There are 1,075 reactions on the combinations of 43 catalysts, five imines, and five thiols.
    \item The asymmetric hydrogenation of olefins (AHO) dataset~\cite{rea_AHO} contains 3,147 products. There are 10,268 reactions with 1,686 transition metal catalysts and 2,754 olefin substrates.
    \item The metabolites and biochemical reactions (BioChem) dataset~\cite{rea_biochem} comprises 16,838 products specifically curated for the biosynthetic planning of natural products, including a total of 33,687 reactions. It does not contain information about reaction conditions.
    \item The USPTO-100 dataset is part of ChemLLMBench~\cite{rea_USPTO_100}, containing 100 products (one product for one reaction) randomly sampled from the respective test sets. It does not contain information about reaction conditions.
\end{itemize}

We carefully examine the overlap between the pre-training dataset and the retrosynthesis benchmark. Except for the USPTO-100 dataset, which is a subset of the USPTO dataset with all 100 reactions included in the pre-training dataset, the other datasets have little or no overlap with the pre-training dataset. Without considering reaction conditions, the ELNs BH, AHO, and BioChem datasets contain 3, 77, and 172 reactions from the pre-training dataset, respectively. Considering the reaction conditions, only the BioChem dataset has 53 overlapping reactions with the pre-training dataset. We also check the overlap within the retrosynthesis benchmark itself, finding that only the AAAA and AHO datasets have one and eight reactions from the BioChem dataset, respectively. Given the minimal overlap among the datasets, we believe it is sufficient to estimate the retrosynthesis prediction ability of models under zero-shot conditions without additional processing.
\begin{figure*}[h]
\centering
    \includegraphics[width=0.75\linewidth]{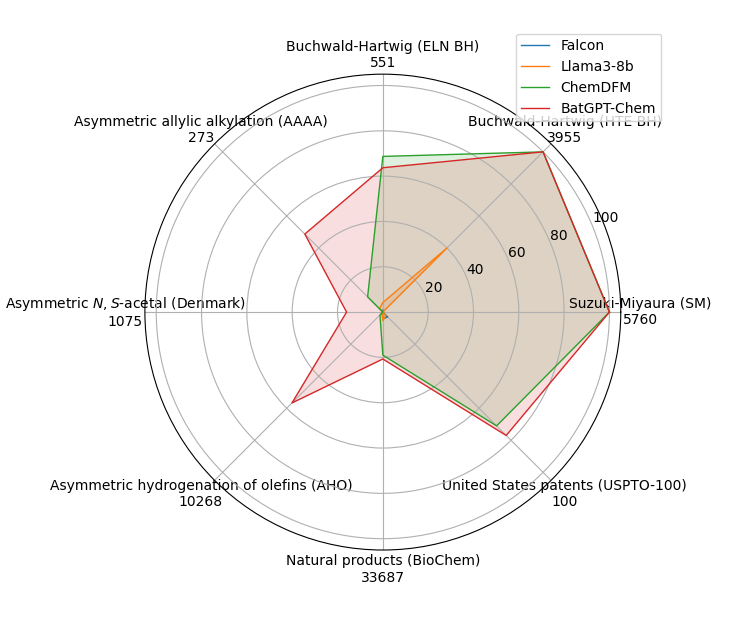}
    \caption{Top-10 MaxFrag accuracy of prediction of different datasets.}
    \label{fig:maxfrag}
\end{figure*}

Fig.~\ref{fig:tmap_0} presents annotated reaction graphs created using the DRFP~\cite{drfp} fingerprints, with colors corresponding to the eight data sources. Despite slight fragmentation in the TMAP sub-trees due to the large disparity in reaction numbers from different sources, related reaction types are well grouped. For example, in the lower left, the Suzuki-Miyaura reactions are clustered continuously at the edges of the spherical tree, while the $N,S$-acetal reactions form disjointed sub-trees in the upper region. The various reaction types span a vast chemical reaction space, posing a significant challenge to the model's generalization performance. Here we introduce two new metrics, coverage of reactants (coverage) and intersection of reaction conditions (intersection), along with MaxFrag and Validity, to thoroughly evaluate retrosynthetic models. 
\newline \textbf{Coverage} of reactants, which indicates whether the true reactant molecules are covered by the model outputs, reflecting the accuracy of the reaction predictions. It differs from the coverage of precursors~\cite{TF_6}, which is related to whether the model can predict at least one valid precursors suggestion at the reaction level. 
\newline \textbf{Intersection} of reaction conditions, which measures whether the model predicts any of the true reaction condition molecules. Directly predicting exact reaction conditions is extremely challenging, leading to common metrics such as Exact Match and Coverage being close to zero for most models. To address this, we propose the intersection accuracy metric, which relaxes the requirements and provides a more attainable measure of model performance when considering reaction conditions.
\newline \textbf{MaxFrag}~\cite{maxfrag}, which accesses the ability to identify principal transformation for classical retrosynthesis. Prediction of the largest fragment focusing only on main compound transformations is the minimal information required for an efficient retrosynthesis route. 
\newline \textbf{Validity}, which measures how many of the SMILES codes predicted by the model are legal and do not violate chemical principles. 

We also conduct multiple experiments by adjusting the Top-$k$ hyperparameter of the model inference to evaluate its performance further. Note that the prediction is correct for products with multiple synthetic pathways as long as the model successfully predicts any of them. 


\begin{table*}[!htp]
\setlength\tabcolsep{9pt}
    \tiny
    \centering
    \scriptsize
    \begin{tabular}{lrrrr}
         \toprule
         \midrule
         Dataset & Falcon & Llama3-8b & ChemDFM & BatGPT-Chem \\
         \midrule
(a) SM & 0.0 & 0.0 & $\mathbf{100.0}$ & $\mathbf{100.0}$ \\
(b) HTE BH & 0.0 & 0.0 & $\mathbf{100.0}$ & $\mathbf{100.0}$ \\
(c) ELN BH & 0.0 & 0.0 & $\mathbf{61.2}$ & 60.8 \\
(d) AAAA & 0.0 & 0.5 & 0.0 & $\mathbf{36.5}$ \\
(e) Denmark & 0.0 & 0.0 & 0.0 & $\mathbf{16.0}$ \\
(f) AHO & 0.4 & 0.4 & 1.8 & $\mathbf{56.7}$ \\
(g) BioChem & 2.7 & 3.4 & 17.8 & $\mathbf{19.9}$ \\
(h) USPTO-100 & 0.0 & 0.0 & 65.0 & $\mathbf{70.0}$ \\
        \bottomrule
    \end{tabular}
    \caption{Top-10 Coverage of reactions for zero-shot retrosynthesis prediction benchmark.}
    \label{tab:res_coverage}
\end{table*}

\subsection{BatGPT-Chem achieves exceptional accuracy in identifying reactants}\label{sec:reactant}
In the MaxFrag score analysis presented in Fig.~\ref{fig:maxfrag}, BatGPT-Chem achieves state-of-the-art performance across most datasets, except for the ELN BH dataset where it is surpassed by ChemDFM. Falcon and Llama3-8b display significantly weaker performance across all datasets, with Llama3-8b exceeding 10\% accuracy only on the HTE BH dataset (achieving 40\%). For the complex reaction class of asymmetric $N,S$-acetal formation from the Denmark dataset, other models fail to identify the largest fragments in any of the 25 reactions, while BatGPT-Chem successfully predicts key fragments in four reactions.

Notably, in reactions from the AHO dataset, which are relatively simple and primarily involve the addition of hydrogens to reactants to obtain the product, other models exhibit extremely poor performance. In contrast, BatGPT-Chem maintains a high accuracy rate. Among the 10,268 reactions in the AHO dataset, it achieves an accuracy of 56.7\%. This accuracy notably exceeds that of ChemDFM by over 30 times, while Falcon and Llama3-8b manage only a 0.4\% accuracy rate.

In direct synthesis scenarios that require precise identification of all interacting components, we evaluate the predictive performance of models by assessing their ability to cover all reactants in a reaction, i.e., coverage. The results, presented in Table~\ref{tab:res_coverage}, align with trends seen in the MaxFrag metric, yet with notably lower coverage scores. For instance, BatGPT-Chem’s performance on the ELN BH and USPTO-100 datasets declines from 63.7\% to 60.8\% and 77.0\% to 70.0\%, respectively. Coverage score of ChemDFM even drops to 0\% on the SM and AAAA datasets. While classical retrosynthesis primarily targets key transformations, the ability to accurately predict additional reactants enhances the completeness of retrosynthetic analyses.

In this evaluation, we assess language models (LLMs) on reactant prediction using the MaxFrag and Coverage metrics instead of the traditional exact match Top-$n$ accuracy. Many LLMs do not differentiate between reactants and reaction conditions during training, typically using the '.' notation to segregate molecules within reaction SMILES strings. This convention complicates the automatic extraction of condition components from reaction SMILES, making it difficult to discern reactants from the output strings, and thus potentially rendering the application of the exact match metric unfair.

\subsection{BatGPT-Chem capably predicts reaction conditions explicitly}

Beyond the identification of reaction reactants, the prediction of appropriate reaction conditions, like catalysts and solvents, remains a crucial and challenging subsequent task~\cite{TF_6}. 
Due to the significant complexity involved in accurately and comprehensively predicting reaction conditions, LLMs often struggle to yield reasonable results when assessed using Exact Match and MaxFrag metrics. Therefore, here we employ the `Intersection' metric, as described in Section~\ref{subsec:data}, to more effectively evaluate the model's capability in predicting reaction conditions.

\begin{figure*}[h]
\centering
    \includegraphics[width=0.75\linewidth]{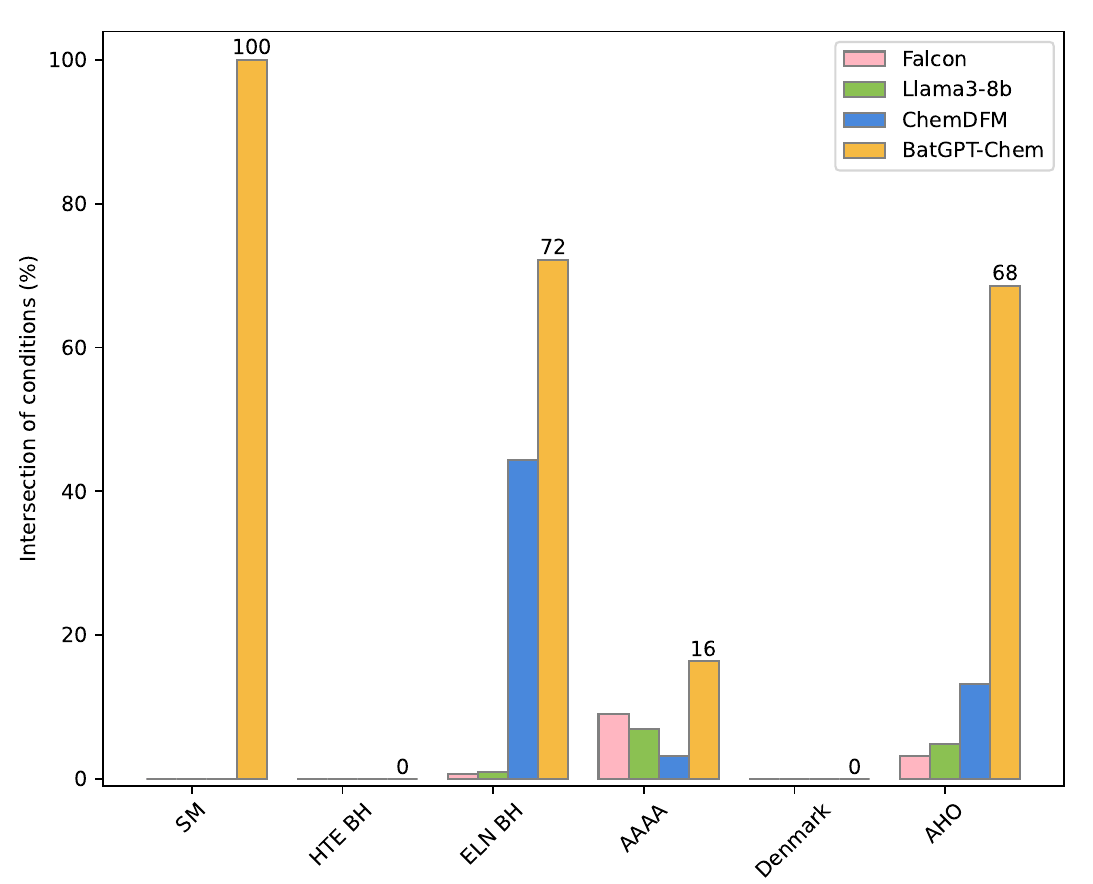}
    \caption{Top-10 Intersection accuracy of prediction of different datasets.}
    \label{fig:intersection}
\end{figure*}

We assess the accuracy of reaction condition predictions across six datasets which contain reaction condition information. As shown in Figure~\ref{fig:intersection}, BatGPT-Chem outperforms other methods by large margins in terms of Intersection rate. 
This demonstrates BatGPT-Chem's exceptional ability to predict reaction conditions and complete retrosynthesis routes effectively. However, it is important to note that on the HTE BH and Denmark datasets, all models consistently fail to present any feasible reaction conditions, highlighting the inherent challenges of this task.

Moreover, as mentioned in Section \ref{sec:reactant}, while many methods do not distinguish between reactants and reaction conditions within their reaction strings, BatGPT-chem's training corpus separates these two elements with a `$>$'. This distinction enables BatGPT-chem to explicitly predict reaction conditions through prompting, markedly enhancing its capacity to provide comprehensive retrosynthesis routes. 

\begin{figure*}[htbp]
\centering
    \includegraphics[width=0.85\linewidth]{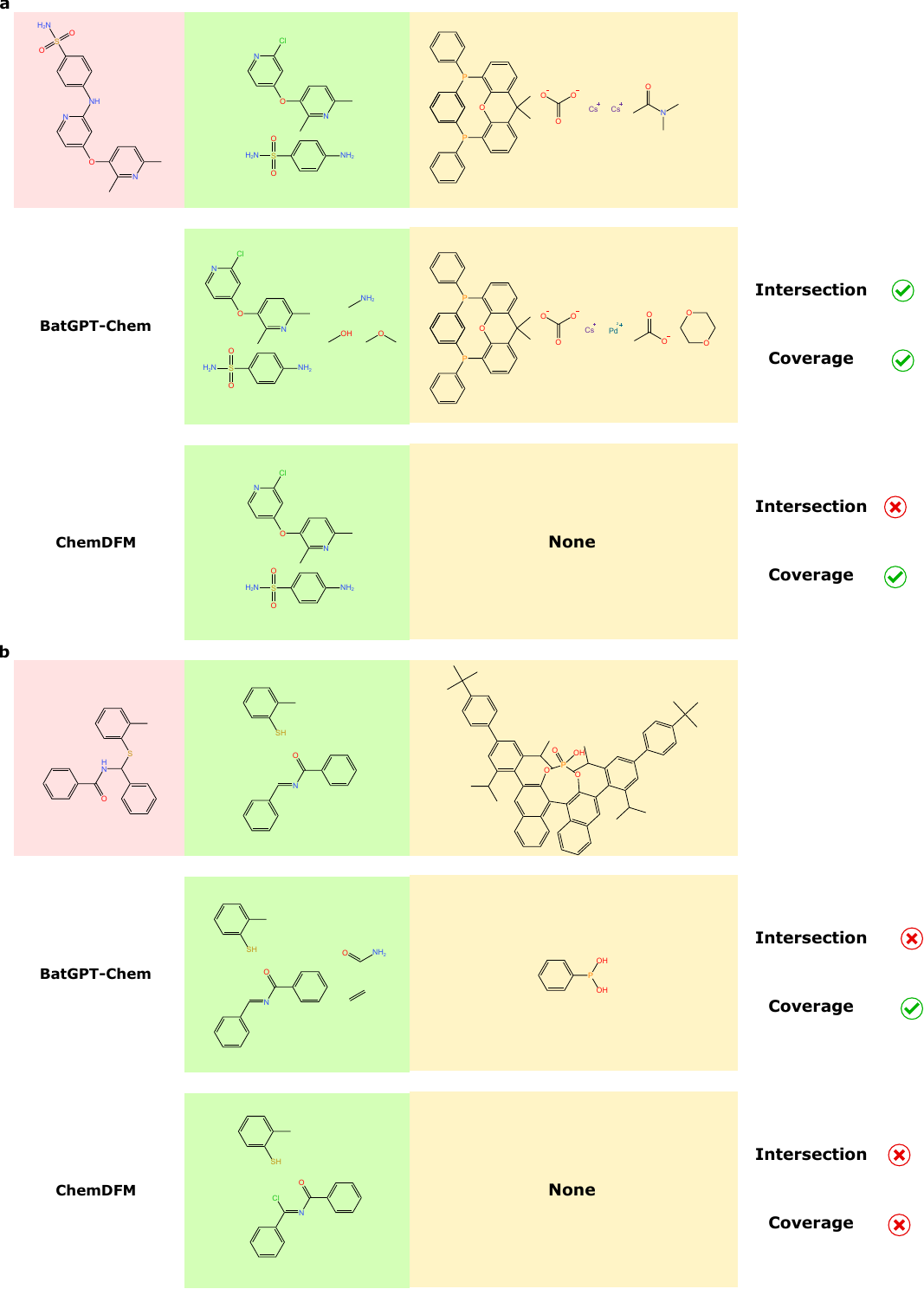}
    \caption{Comparison of predictions between BatGPT-Chem and ChemDFM where products are displayed in pink blocks, reactants are in green blocks, and reaction conditions are in yellow blocks. $\mathbf{a}$ An example from the ELN BH dataset. $\mathbf{b}$ An example from the Denmark dataset.}
    \label{fig:case}
\end{figure*}

To further illustrate the predictive capabilities concerning reaction conditions, we analyze two specific cases: one from the ELN BH dataset and another from the Denmark dataset, comparing BatGPT-Chem with the top-performing baseline, ChemDFM. 
As shown in Figure~\ref{fig:case}a, BatGPT-Chem successfully predicts most conditions, including the catalyst and metal, only missing one reaction condition. In contrast, ChemDFM fails to generate any correct conditions. In this reaction, the catalyst's structure is relatively complex, and multiple reaction conditions are required for this retrosynthesis pathway. In cases where reaction conditions are exceedingly complex, it becomes even more challenging for models to make accurate predictions. As can be seen in Figure~\ref{fig:case}b, BatGPT-Chem covers all reactants but fails to predict the catalyst, whereas ChemDFM correctly identifies only one reactant. Note that BatGPT-Chem occasionally recommends supplementary simple small molecules among the reactants, which have a negligible influence on the determination of the reaction pathway.

\begin{figure*}[htbp]
\centering
    \includegraphics[width=0.95\linewidth]{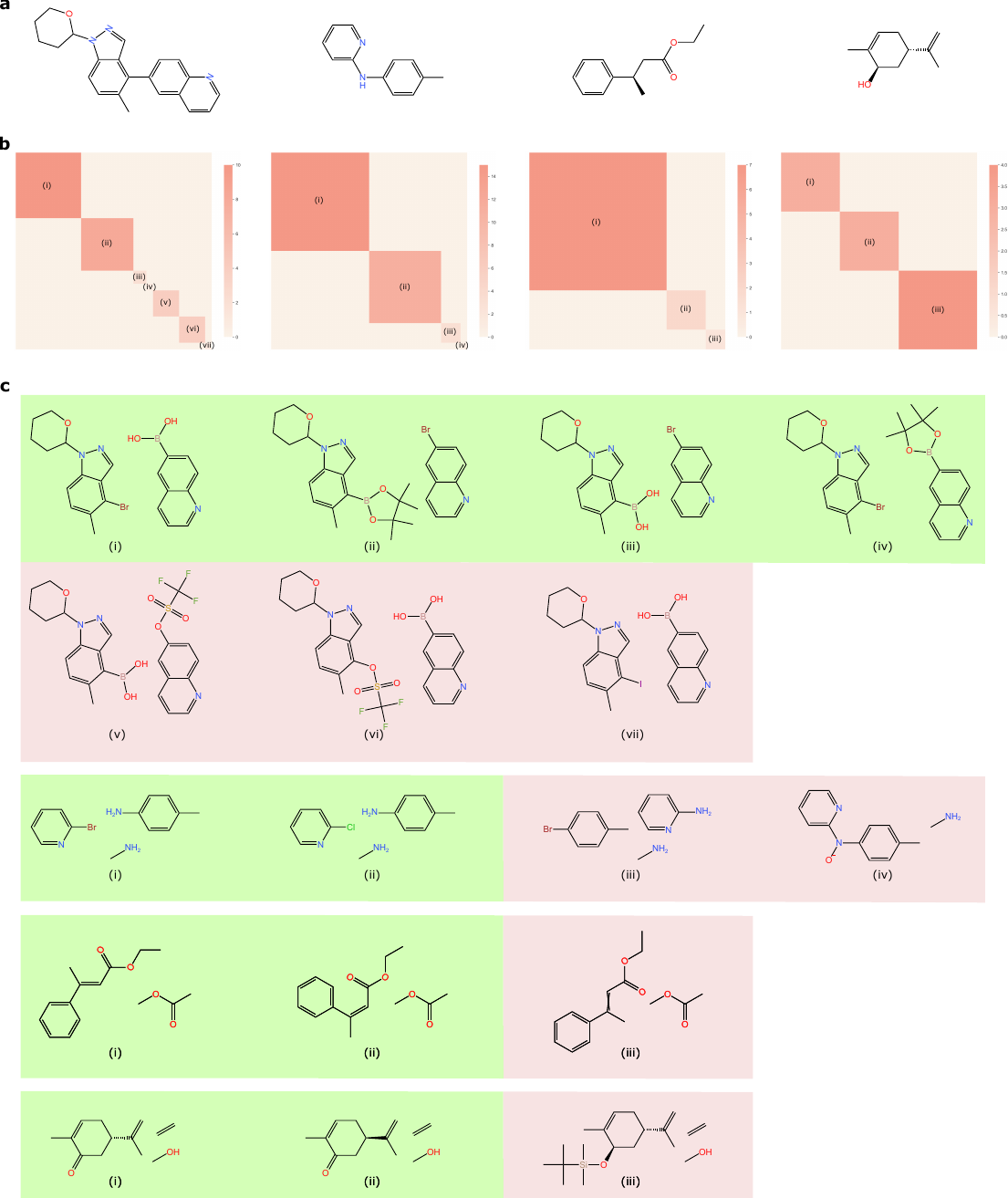}
    \caption{Analysis of predictions generated by BatGPT-Chem. $\mathbf{a}$ Products sampled from the SM, the HTE BH, the AHO, and the BioChem dataset, respectively. $\mathbf{b}$ Numbers of prediction within Top-$k$. $\mathbf{c}$ Details of predictions where green means ground truth is covered and red means not.}
    \label{fig:multi}
\end{figure*}
\begin{table*}[!htp]
\setlength\tabcolsep{9pt}
    \tiny
    \centering
    \scriptsize
    \begin{tabular}{lrrrr}
         \toprule
         \midrule
         Model & Falcon & Llama3-8b & ChemDFM & BatGPT-Chem \\
         \midrule
(a) SM dataset & 0.0 & 40.0 & $\mathbf{100.0}$ & $\mathbf{100.0}$ \\
(b) HTE BH dataset & 40.0 & 76.0 & $\mathbf{100.0}$ & $\mathbf{100.0}$ \\
(c) ELN BH dataset & 45.9 & 77.5 & 99.4 & $\mathbf{100.0}$ \\
(d) AAA dataset & 86.5 & 85.1 & 99.1 & $\mathbf{100.0}$ \\
(e) Denmark dataset & 9.6 & 88.8 & 99.2 & $\mathbf{100.0}$ \\
(f) AHO dataset & 69.1 & 84.1 & 99.6 & $\mathbf{100.0}$ \\
(g) BioChem dataset & 48.7 & 74.7 & 95.1 & $\mathbf{98.5}$ \\
(h) USPTO-100 dataset & 35.1 & 75.3 & 99.3 & $\mathbf{100.0}$ \\
        \bottomrule
    \end{tabular}
    \caption{Top-10 Validity of reactions for zero-shot retrosynthesis prediction benchmark.}
    \label{tab:res_validity}
\end{table*}
\subsection{BatGPT-Chem excels in generating multiple viable retrosynthesis routes}
After achieving accurate predictions for reactants and reaction conditions, the focus shifts to evaluating the model's ability to generate diverse, correct, and even novel predictions. Due to the stochastic nature of LLMs, retrosynthesis predictions sampled from BatGPT-Chem for a fixed product will not be unique. To assess BatGPT-Chem's capability to propose multiple retrosynthesis routes, we examine four representative products (Fig.~\ref{fig:multi}a) from the benchmark. We sample the Top-30 predictions for the SM and HTE BH datasets and the Top-10 predictions for the AHO and BioChem datasets.
Fig.~\ref{fig:multi}b and ~\ref{fig:multi}c display the frequencies and detailed predictions, showcasing BatGPT-Chem's capacity to provide diverse retrosynthesis routes.

Specifically, for the product from the SM dataset, BatGPT-Chem generates seven different predictions, four of which (i to iv) match the ground truths with a frequency ratio of 23/30. The remaining predictions are not in the dataset but still contain valid functional groups. For the product from the HTE BH dataset, BatGPT-Chem provides four different predictions, two of which (i and ii) match the ground truths with a frequency ratio of 26/30. Notably, prediction iii, not included in the original dataset, is found in Reaxys~\cite{BH_case_0, BH_case_1} (ID: 39015457), highlighting the model's capability to predict reactions outside the benchmark dataset. 

For the product from the AHO dataset, BatGPT-Chem produces three different predictions, two of which (i and ii) match the ground truths with a frequency ratio of 9/10. For this chiral product, BatGPT-Chem successfully predicts all cis-trans isomerism (i and ii) and even provides the SMILES without cis-trans information (iii). For the product from the BioChem dataset, BatGPT-Chem makes four different predictions, two of which (i and ii) match the ground truths with a frequency ratio of 6/10. This is also a chiral product, and BatGPT-Chem accurately predicts different chiral configurations $S$ (i) and $R$ (ii). Again, BatGPT gives some extra simple small molecules in the reactants for some predictions. 

These results emphasize BatGPT-Chem's ability to capture the major backbone of the reactant molecules, provide diverse predictions for retrosynthesis pathways, and predict reactions beyond the benchmark dataset, further demonstrating its robustness and versatility in chemical synthesis planning.


In retrosynthesis, the feasibility of synthesizing a single product from various precursors adds complexity to the evaluation of model predictions. For example, in the reaction R-R1+NH3-$>$R-NH2, multiples substituents for R1, such as -OH, -Cl, -Br, -I, or -F are valid, each leading to correct predictions~\cite{maxfrag}. 
The primary distinction among these options stems from their respective reaction rates and yields.
While the `MaxFrag' metric does not fully address this issue, it still represents an effort to better manage such data ambiguities during the validation process. 
Instead of devising a new metric, our approach focuses on addressing this challenge from a data-centric perspective to enhance the comprehensiveness of performance evaluation. Specifically, during dataset compilation for model evaluation, we endeavor to gather as many synthetic pathways for the same product as feasible. Subsequently, during the assessment phase, a prediction is deemed successful if it aligns with any of the collected retrosynthesis pathways for that product. 

\subsection{BatGPT-Chem generates outputs with high validity}

As LLMs sometimes produce SMILES representations of molecules that may not be valid or chemically plausible, we employ RDKit~\cite{rdkit} to verify the validity of molecules generated by these models. BatGPT-Chem consistently achieves high validity rates, nearing or reaching 100\% across all datasets, thereby confirming its strong grasp of chemical language. Given that chemical symbols can be considered a specialized language domain, models primarily trained on general natural language corpora often fail to fully comprehend it. BatGPT-Chem can avoid tedious post-processing of grammatical corrections~\cite{TF_1} to fix the syntax errors of outputs. Furthermore, it successfully interprets the cis-trans and chiral information inherent in chemical language, as illustrated in Fig.~\ref{fig:multi}c.

\section{Discussion}

Large language models (LLMs) have made substantial progress across various fields, demonstrating significant potential to spearhead advancements in AI for Science. Notably, their proficiency in processing sequential data makes them particularly apt for the chemical domain, where common representations such as SMILES also adopt a sequential format. Considering the natural potential of LLMs to learn and predict chemical structures and reactions, we develop BatGPT-Chem, a pioneering large-scale model tailored for retrosynthesis analysis, to address three critical limitations prevalent in existing AI models: i) a deficit in comprehensive molecular and chemical reaction knowledge; ii) oversight of reaction conditions; and iii) inadequate generalization across diverse chemical reactions.

BatGPT-Chem distinguishes itself by its extensive training corpus, which covers a wide range of chemical literature and chemical string data, i.e., SMILES strings. The implementation of carefully crafted prompt templates and tailored instruction-tuning data during pre-training has significantly enhanced its capacity to decipher both natural and chemical languages. This advancement is reflected in its improved accuracy in predicting reactants and the near-perfect validity of its output. Particularly noteworthy is BatGPT-Chem's performance on comprehensive benchmark datasets, demonstrating remarkable zero-shot retrosynthesis prediction capacities that hold practical implications for real-world applications.

A unique aspect of BatGPT-Chem compared to other LLMs is its explicit handling of reaction conditions. By directly extracting reaction conditions from datasets and creating specific prompts to predict them, BatGPT-Chem shows superior ability to elucidate components such as solvents and catalysts. 
When predicting reaction conditions, non-generative models can only deal with a fixed set of molecules and are usually modeled as multi-classification or multi-label problems~\cite{condition_cls_0, condition_cls_1, condition_cls_2, condition_cls_3}, which greatly limits the generalization ability of these models. These methods use a post-processing approach in retrosynthesis, where reaction conditions are predicted after the reactants and products are known. This two-stage processing requires additional training of the model, is not simple enough to use, and makes it more difficult to ensure the stability of the model. 
Contrastingly, generative models like BatGPT-Chem can predict a wide variety of reaction condition molecules, are not constrained by a finite set, and can offer novel and heuristic predictions in an end-to-end manner. Since most of the reaction condition information is stored using raw text~\cite{jiang2021smiles}, constructing datasets for machine learning is inherently time-consuming and laborious, including steps such as extracting text, removing errors, and converting to computer-readable sequences~\cite{reamvp, condition_cls_3}. BatGPT-Chem can provide reference reaction conditions for inverse synthesis datasets like BioChem~\cite{rea_biochem}, which do not contain information on reaction conditions.
This capability can help build more comprehensive and enriched datasets of chemical reactions, facilitating the use of machine learning in chemical reaction modeling.

Moreover, BatGPT-Chem excels in generating diverse and viable retrosynthesis pathways, providing valuable insights for chemists. Actually, the cases where multiple paths correspond to the same product are rare in the pre-training datasets, thus BatGPT-Chem's ability to suggest various feasible routes can be attributed to its profound understanding of chemical reaction mechanisms. By optimizing beam search strategies and temperature settings, the model adeptly balances diversity and correctness. In contrast, efforts to enhance output diversity in other LLMs through elevated temperature settings often result in an increased error rate.

In conclusion, BatGPT-Chem sets new benchmarks for effective and dependable AI-driven retrosynthesis planning. However, the work is still tempered by the quality of the data, primarily from open-access sources. Despite endeavors to enrich reaction condition data and compile comprehensive retrosynthesis pathways, gaps remain. Future improvements will likely require collective efforts from across the scientific community. Another constraint is the scope of chemical languages covered; currently focused on SMILES, exploring additional string-based representations like SELFIES~\cite{SELFIES} could broaden our model's utility, paving the way for its application to a greater spectrum of chemical reactions.

\section{Methods}

\subsection{Unified modeling}
\begin{figure}[h]
\centering
    \includegraphics[width=1.0\linewidth]{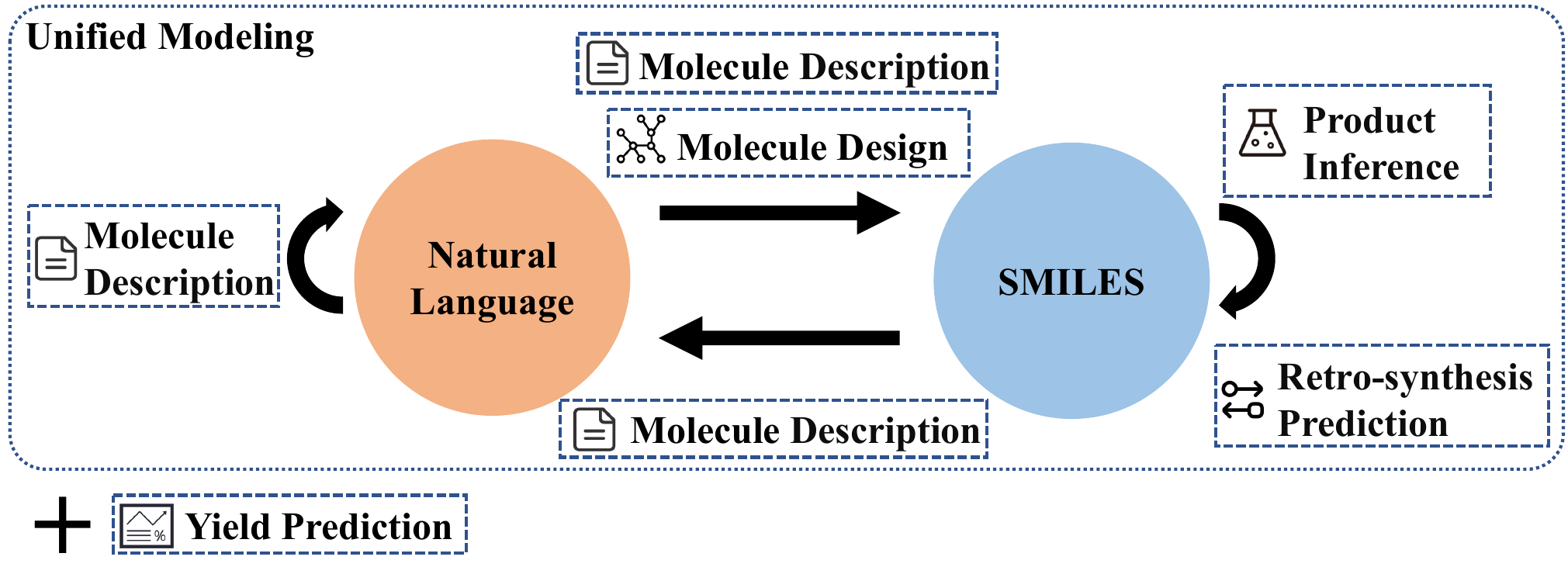}
    \caption{The illustration of our unified modeling between natural language and SMILES.}
    \label{fig:modeling}
\end{figure}

Viewing natural language as a specialized language, we can employ LLMs for unified modeling of natural language to SMILES, SMILES to natural language, SMILES to SMILES, and natural language to natural language. This naturally facilitates the completion of various chemistry tasks: \textbf{Molecule Description}, \textbf{Molecule design}, \textbf{Product Inference}, and \textbf{Retro-synthesis Prediction}. Additionally, we have also modeled the \textbf{Yield Prediction} task. We showcase our modeling approach in Fig.~\ref{fig:modeling}. We model molecule description as bidirectional conversions between natural language and SMILES, as well as conversions between natural language. We model molecule design as a conversion from natural language to SMILES. We also model product inference and retro-synthesis prediction as conversions from SMILES to SMILES. Additionally, we have also included a task for yield prediction.

\subsection{Chemistry tasks and prompt templates}
Following the modeling approach outlined above, we focus on the key tasks in the chemistry domain: \textbf{Molecule Description}, \textbf{Molecule design}, \textbf{Product Inference}, \textbf{Retro-synthesis Prediction}, and \textbf{Yield Prediction}. We construct instruction tuning datasets based on existing chemical, drug, and medicine datasets using prompt templates, to train models capable of addressing these tasks.

\textbf{Retro-synthesis Prediction.} Retro-synthesis Prediction is a crucial task for chemistry. It involves inferring possible reaction pathways and conditions by given product molecules, thereby reverse-predicting the synthetic route to generate the product. Retro-synthesis prediction enables researchers to explore and discover new organic molecular structures more rapidly, which is essential for fields such as organic synthesis chemistry and drug discovery. We train the model's retro-synthesis prediction capability using two subtasks: 1) Reactant and catalyst prediction: Given a product, predict the potential catalysts and reactants that may be required. 2) Reactant prediction: Given a product and catalyst, predict the reactants.

\textbf{Product Inference.} Product inference aims at predicting the products based on given starting materials and specific reaction conditions, which holds significant importance in fields such as organic synthesis and drug design. We train the model's product inference capability using two subtasks: 1) Product and catalyst prediction: Given reactants, predict the potential catalysts and products that may be involved. 2) Product prediction: Given products and catalysts, predict the reactants involved.

\textbf{Molecule Design.} Molecule Design is a field involving the creation of new molecules using theoretical and computational methods to produce molecular structures with specific properties or functionalities. This field plays a crucial role in various domains including chemistry, drug design, and materials science. The aim of molecule design is to systematically generate molecules with desired properties and activities to meet specific application needs. This work fully considers over a hundred molecular properties, such as molecule weight, valence electron count, Balaban J value, BertzCT value, number of heavy atoms, number of NHs or OHs, and number of nitrogen and oxygen atoms. It is hoped that the LLM can take into account researchers' specific requirements for molecule properties of catalysis, products, and reactants of chemical reactions. To train the model's molecular design capability, the following three tasks are adopted: 1) Specifying catalyst molecular properties: Given reactants to produce a specific product, the catalyst is required to meet certain properties. 2) Specifying reactant and catalyst molecular properties: Given the desired product to be synthesized, both reactants and catalysts are required to meet certain properties. 3) Specifying reactant, catalyst, and product properties: The model is required to provide a chemical reaction with specified molecular properties for reactants, catalysts, and products.

\textbf{Molecule Description.} Molecule description refers to using computational models to predict and describe the function, effects, and related properties of a molecule given its name, SMILES, or other representations. We adopt the following eight subtasks to train the model's molecule description capability. We not only utilize chemical data for training to enable the model to fully understand and perceive the correspondence between molecular names in both English and Chinese, molecule descriptions, molecule SMILES, and molecule IUPAC (International Union of Pure and Applied Chemistry chemical nomenclature) names, thus obtaining strong molecular description capabilities, but also incorporate some pharmaceutical data to enhance the model's ability in the pharmaceutical field: 1) Given the molecular Chinese name, generate the English name and description. 2) Given the molecular English name, generate the Chinese name and description. 3) Given the molecular description, generate the Chinese name and English name of the molecule. 4) Given the molecular description, generate the IUPAC name and SMILES code of the molecule. 5) Given the molecular SMILES code, generate the IUPAC name and SMILES code of the molecule. 6) Given the molecular IUPAC name, generate the SMILES code and description of the molecule. 7) Given the Chinese name of a drug, generate the English name and description. 8) Given the English name of a drug, generate the Chinese name and description. 9) Given the description of a drug, generate the Chinese name and English name.

\textbf{Yield Prediction.} Yield prediction in chemical reactions refers to the estimation, through experimental or computational methods, of the ratio between the actual quantity of products generated in a chemical reaction and the theoretically maximum possible yield. We also train the model by predicting corresponding yields for given chemical reactions.

\subsection{Pre-training data source}
We utilize publicly available high-quality datasets in the field of chemistry, as well as close-source datasets within our own team, as the raw datasets. Then, we transform them into instruction tuning datasets using the aforementioned prompt templates.

\subsubsection{Publicly Available Datasets}
\begin{itemize}
    \item \textbf{USPTO}~\cite{lowe2012extraction} USPTO collects reaction data extracted through text mining from United States patents published between 1976 and September 2016.
    \item \textbf{CHEBI}~\cite{degtyarenko2007chebi} Chemical Entities of Biological Interest (CHEBI) is a freely available dictionary of molecular entities focused on ``small'' chemical compounds. The term ``molecular entity'' refers to any constitutionally or isotopically distinct atom, molecule, ion, ion pair, radical, radical ion, complex, conformer, etc., identifiable as a separately distinguishable entity. The molecular entities in question are either products of nature or synthetic products used to intervene in the processes of living organisms.
    \item  \textbf{CJHIF}~\cite{jiang2021smiles} Chemical Journals with High Impact factors (CJHIF) is a high-quality dataset containing a large number of chemical reaction equations extracted from various chemical journals.
    \item \textbf{PubChem}~\cite{kim2016pubchem} PubChem is an open chemistry database at the National Institutes of Health (NIH), which mostly contains small molecules, but also larger molecules such as nucleotides, carbohydrates, lipids, peptides, and chemically-modified macromolecules.
    \item \textbf{Text2Mol}~\cite{edwards-etal-2021-text2mol} Text2Mol provides a large amount of data containing natural language descriptions of molecules.
\end{itemize}

\subsubsection{Close Source Datasets}
\begin{itemize}
    \item \textbf{Drug Instruction} We collect a large number of drug names, drug descriptions, and corresponding molecular formulas from drug instruction to enhance the model's capabilities in the pharmaceutical domain.
    \item \textbf{Organic Compound Manual} We have a large collection of private organic compound manuals, containing information such as organic compound names, compound descriptions, compound SMILES, etc.
    \item \textbf{Molecular Formula and Name Reference Table} We have collected a large amount of publicly available data on compound names and their corresponding molecular formulas.
    \item \textbf{SMILES, IUPAC Names, and Molecular Descriptions Reference Table} We have collected data on SMILES, IUPAC names, and their corresponding molecular descriptions.
\end{itemize}

\subsection{Data Transformation For Instruction Tuning}
We extract reaction data into reactant SMILES, catalyst SMILES, product SMILES, and yield data. Then we conduct data augmentation, that is if there are multiple reactants, catalysts, or products, we shuffle the SMILES of these compounds.

For retro-synthetic prediction, product inference, and yield inference, we organize reactant SMILES, catalyst SMILES, product SMILES, and yield data according to the prompt templates. For molecule design, we use the RDKit tool to randomly select 1-20 properties from a candidate pool of 172 properties to fill in the prompt templates. 

\subsection{Data Details}


\begin{table}[!htp]
\setlength\tabcolsep{12pt}
    \tiny
    \centering
    \scriptsize
    \begin{tabular}{lc}
         \toprule
         \midrule
         \bf Task & Amount\\
         \midrule
         Retro-synthesis Prediction & 30114006 \\
         Product Inference & 30114006  \\
         Molecule Design & 40695857 \\
         Molecule Description & 210469 \\
         Yield Prediction & 10775991 \\
         \midrule
         Total & 111910329 \\
         \midrule
         \bottomrule
    \end{tabular}
    \caption{Data details.}
    \label{tab:data details}
\end{table}

Table~\ref{tab:data details} lists the data scale used for each task. We have over a hundred million data entries in total, with an average length exceeding 150 tokens. The total number of tokens trained exceeds 15 billion.

\subsection{Training Details}
\subsubsection{Base Model}
We select our team's self-developed BatGPT-15B model~\cite{li2023batgpt} as the base model for instruction tuning. BatGPT-15B is a large bilingual model for both Chinese and English, pre-trained using bidirectional autoregressive methods, and has demonstrated excellent performance on public benchmarks such as CMMLU~\cite{li2023cmmlu}.

\subsubsection{Vocabulary Expansion}
Since the BatGPT-15B model is originally designed for natural language, particularly Chinese and English, it lacks comprehensive coverage of specialized terms in chemistry or SMILES. Consequently, expanding its vocabulary becomes necessary. We employ the Byte Pair Encoding (BPE) algorithm to train a vocabulary using diverse training data, encompassing various forms of molecular SMILES, chemical equation SMILES expressions, molecular names, and more. We also include all chemical element symbols in the augmented vocabulary to empower the model with the potential to handle all chemical elements. Subsequently, we merge this augmented vocabulary with that of BatGPT-15B, ultimately yielding a final vocabulary size of 151851.

\subsubsection{Training Settings}
We train our model using the deepspeed zero2 strategy on an Nvidia A800 GPU cluster. We set the maximum length to 2048, the batch size per GPU to 8, utilize the AdamW optimizer with a learning rate of 2e-4, and employ the cosine learning rate schedule strategy. We enable gradient checkpointing and set max gradient normalization to 1.0 and weight decay to 0.1.

\section{Data and Code Availability}
Working in progress.

\section{Acknowledgements}
Working in progress.

\section{Authors' Contributions}
Working in progress.

\section{Competing Interests}
The authors declare that they have no competing interests.

\backmatter

\bibliography{sn-bibliography}

\begin{appendices}
\section{Case Study}\label{app:case}
\begin{figure*}[h]
\centering
    \includegraphics[width=0.65\linewidth]{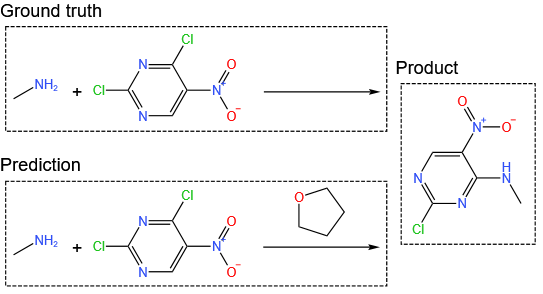}
    \caption{Case 1 from USPTO-50k. Give the product \textbf{CNc1nc(Cl)ncc1[N+](=O)[O-]}, the model successfully predicts the correct reactant \textbf{CN.O=[N+]([O-])c1cnc(Cl)nc1Cl}. It also simultaneously provides a potential catalyst \textbf{C1COCC1}, which is a commonly used catalyst.}
    \label{fig:uspto0}
\end{figure*}

\begin{figure*}[h]
\centering
    \includegraphics[width=0.95\linewidth]{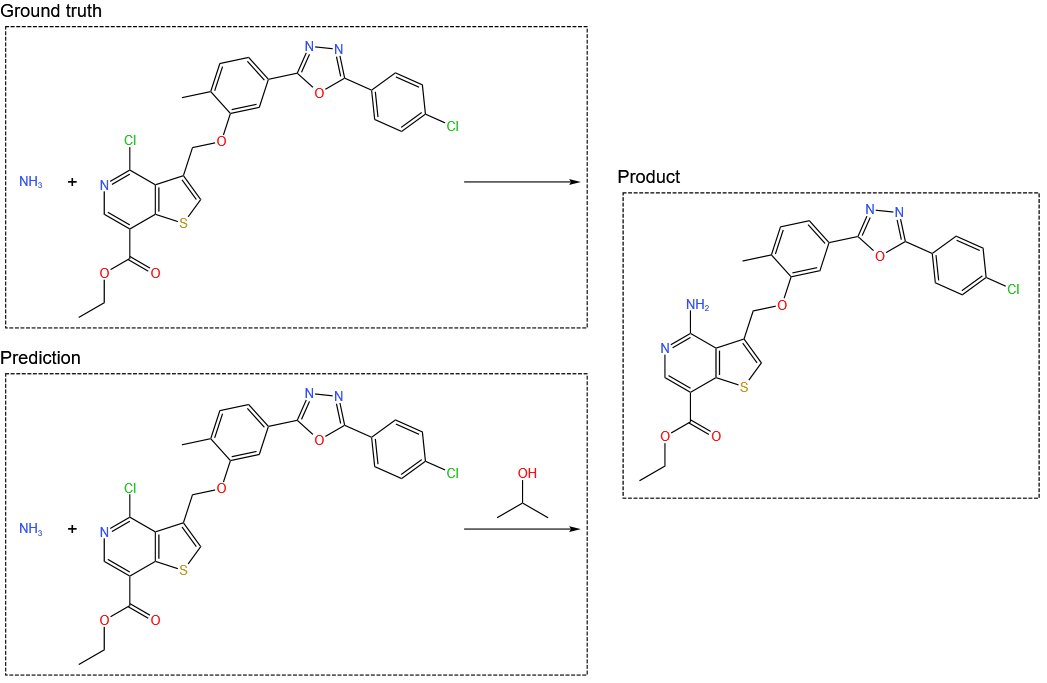}
    \caption{Case 2 from USPTO-50k. Give the product \textbf{CCOC(=O)c1cnc(N)c2c(COc3cc(-c4nnc(-c5ccc(Cl)cc5)o4)ccc3C)csc12}, the model successfully predicts the correct reactant \textbf{N.Clc1c2c(scc2COc2c(C)ccc(-c3nnc(-c4ccc(Cl)cc4)o3)c2)c(C(OCC)=O)cn1}. The model also predicts a catalyst \textbf{C(C)(O)C}, which could act as a solvent.}
    \label{fig:USPTO_2}
\end{figure*}

\begin{figure*}[h]
\centering
    \includegraphics[width=0.95\linewidth]{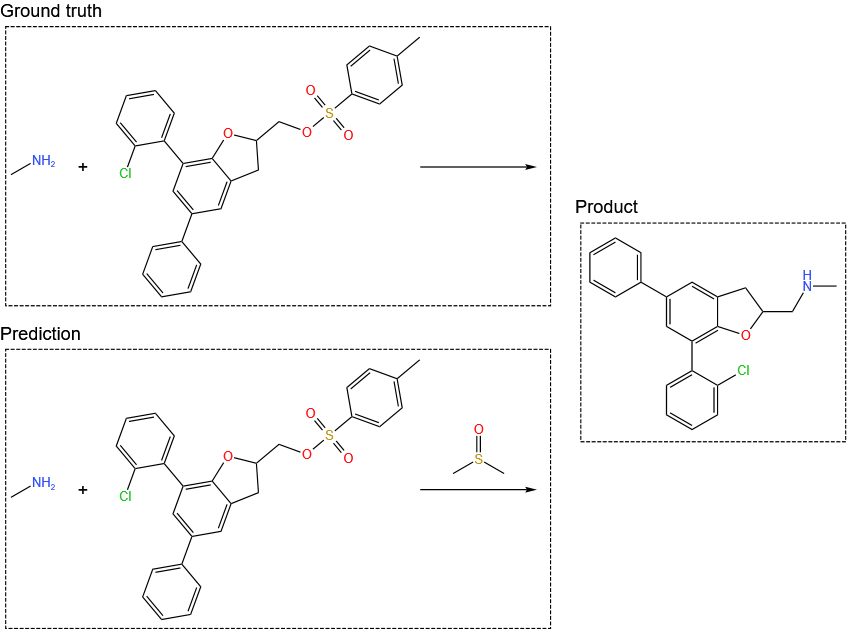}
    \caption{Case 3 from BioChem. 
Give the product \textbf{CNCC1Cc2cc(-c3ccccc3)cc(-c3ccccc3Cl)c2O1}, the model predicts the correct reactant \textbf{CN.Cc1ccc(S(=O)(=O)OCC2Cc3cc(-c4ccccc4)cc(-c4ccccc4Cl)c3O2)cc1}. A \textbf{catalyst S(C)(=O)C}, which bears a resemblance to the reactant structure, is predicted, possibly serving as a "reaction fragment" or an "intermediate product."}
    \label{fig:BioChem_0}
\end{figure*}

\begin{figure*}[h]
\centering
    \includegraphics[width=0.95\linewidth]{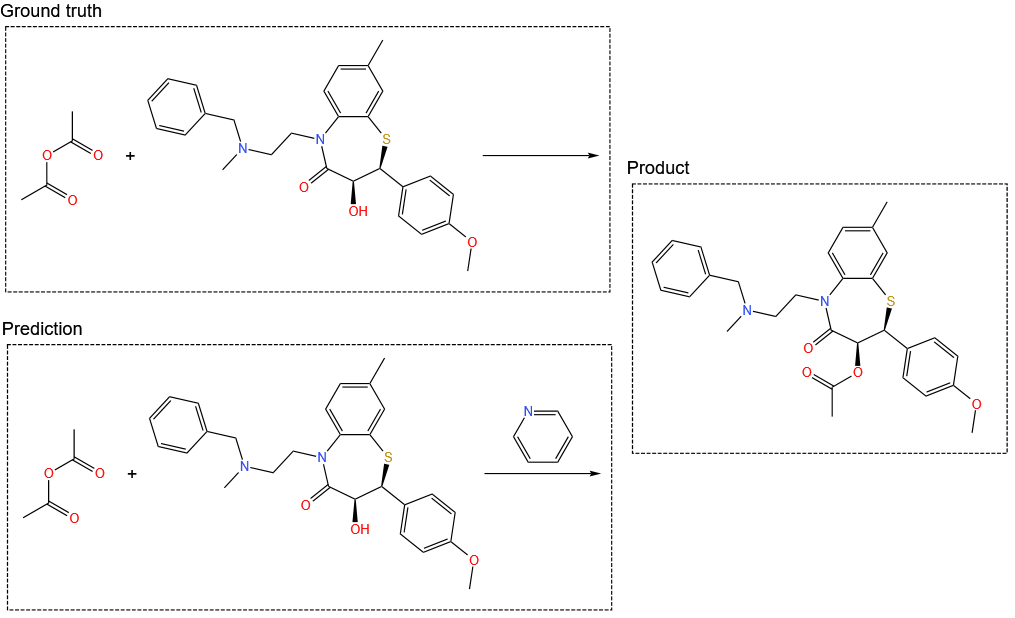}
    \caption{Case 4 from BioChem. For the product \textbf{COc1ccc([C@@H]2Sc3cc(C)ccc3N(CCN(C)Cc3ccccc3)C(=O)[C@@H]2OC(C)=O)cc1}, the model successfully predicts the reactant \\ \textbf{CC(=O)OC(C)=O.c12ccc(C)cc1S[C@@H](c1ccc(OC)cc1)[C@@H](O)C(=O)N2CCN(C)Cc1ccccc1} and provides a catalyst \textbf{c1cccnc1}, which could be a solvent.}
    \label{fig:}
\end{figure*}

\begin{figure*}[h]
\centering
    \includegraphics[width=0.7\linewidth]{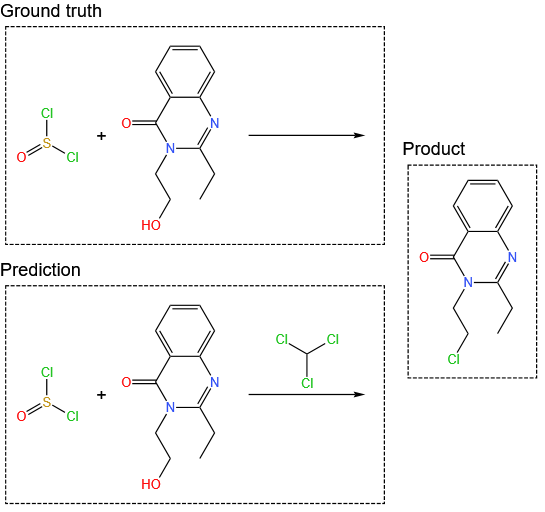}
    \caption{Case 5 from BioChem. For the product \textbf{CCc1nc2ccccc2c(=O)n1CCCl}, the model predicts the correct reactants \textbf{O=S(Cl)Cl.c12ccccc1nc(CC)n(CCO)c2=O}, and provides a catalyst \textbf{ClC(Cl)Cl}, which could be a solvent.}
    \label{fig:BioChem_2}
\end{figure*}

\begin{figure*}[h]
\centering
    \includegraphics[width=0.95\linewidth]{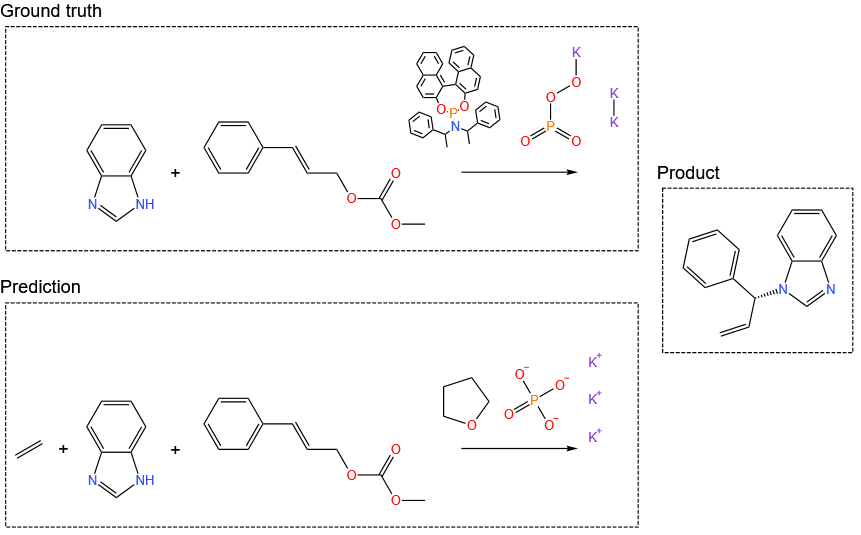}
    \caption{Case 6 from our private benchmark AAAA. For the product \textbf{C=C[C@H](c1ccccc1)n1cnc2ccccc21}, the model predicts a reactant \textbf{C=C.c12ccccc1[nH]cn2.O=C(OC/C=C/c1ccccc1)OC} with an additional small molecule ethylene, and successfully predicts furan as the solvent. The solvent information appears only in the original paper of this reaction, demonstrating that the model successfully transfers knowledge from the literature to retro-synthesis prediction tasks after being trained on a large dataset.}
    \label{fig:AAA_0}
\end{figure*}

\begin{figure*}[h]
\centering
    \includegraphics[width=0.95\linewidth]{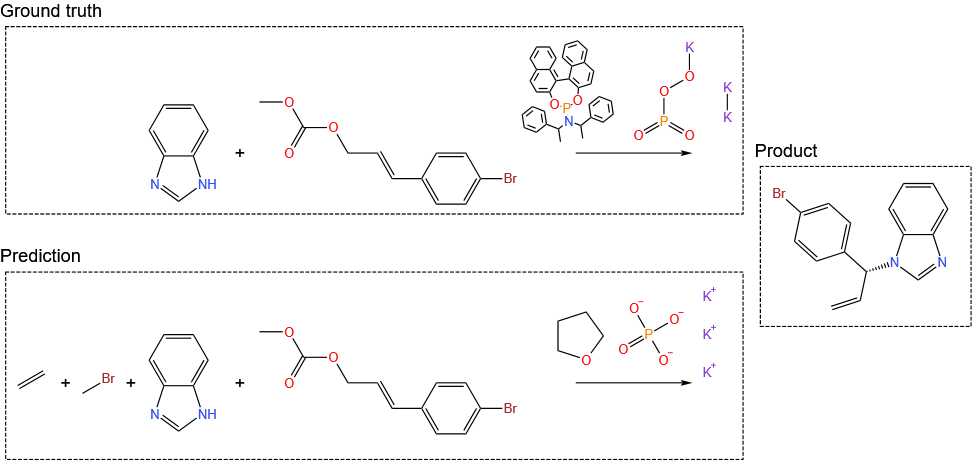}
    \caption{Case 7 from our private benchmark AAAA. For the reactant \textbf{C=C[C@H](c1ccc(Br)cc1)n1cnc2ccccc21}, the model correctly predicts the reactant \textbf{c1nc2ccccc2[nH]1.CBr.C=C.COC(=O)OC\textbackslash\textbackslash C=C\textbackslash\textbackslash c1ccc(Br)cc1} but includes an additional ethylene and its corresponding hydrogen halide molecule. The main reason might be the presence of a halogen substituent on the reactant's ring, which is relatively reasonable. The  ligand is not predicted, but furan is successfully predicted as the solvent.}
    \label{fig:AAA_1}
\end{figure*}




\end{appendices}


\end{document}